\pdfoutput=1
\def\year{2021}\relax
\documentclass[letterpaper]{article} 
\usepackage{aaai21}  
\nocopyright
\usepackage{times}  
\usepackage{helvet} 
\usepackage{courier}  
\usepackage[hyphens]{url}  
\usepackage{graphicx} 
\usepackage{natbib}  
\usepackage{caption} 
\title{Debona: Decoupled Boundary Network Analysis\\for Tighter Bounds and Faster Adversarial Robustness Proofs}
\author{%
	Christopher Brix, Thomas Noll \\
	Software Modeling and Verification Group \\
	RWTH Aachen University \\
	Aachen, D-52074 \\
	\texttt{christopher.brix@rwth-aachen.de}, \texttt{noll@cs.rwth-aachen.de}
	\vspace{-3mm}
}
\affiliations{

}

\usepackage{csquotes}
\usepackage{amsmath}
\usepackage{subcaption}
\usepackage{pgfplots}
\usepackage{booktabs}       
\DeclareMathOperator*{\argmax}{arg\,max}
\DeclareMathOperator*{\argmin}{arg\,min}
\newcommand\restr[2]{{
		\left.\kern-\nulldelimiterspace 
		#1 
		\vphantom{\big|} 
		\right|_{#2} 
	}}
	
	\usepackage[toc,page]{appendix}

\begin{document}

\maketitle

\begin{abstract}
  Neural networks are commonly used in safety-critical real-world applications.
  Unfortunately, the predicted output is often highly sensitive to small, and possibly imperceptible, changes to the input data.
  Proving that either no such adversarial examples exist, or providing a concrete instance, is therefore crucial to ensure safe applications.
  As enumerating and testing all potential adversarial examples is computationally infeasible,
  verification techniques have been developed to provide mathematically sound proofs of their absence using overestimations of the network activations.
  We propose an improved technique for computing tight upper and lower bounds of these node values, based on increased flexibility gained by computing both bounds independently of each other.
  Furthermore, we gain an additional improvement by re-implementing part of the original state-of-the-art software \enquote{Neurify}, leading to a faster analysis.
  Combined, these adaptations reduce the necessary runtime by up to 94\%, and allow a successful search for networks and inputs that were previously too complex. 
  We provide proofs for tight upper and lower bounds on max-pooling layers in convolutional networks.
  To ensure widespread usability, we open source our implementation \enquote{Debona}, featuring both the implementation specific enhancements as well as the refined boundary computation for faster and more exact~results.
\end{abstract}

\noindent \section{Introduction}
\label{sec:introduction}

Major advances in both the theory and implementation of neural networks have enabled their employment in many application domains including safety-critical ones, such as autonomous driving \citep{bojarski2016end}, healthcare \citep{esteva2019guide}, or military \citep{wu2015typical}.
However, experience shows that neural networks often lack \emph{robustness} properties, meaning that small, or even imperceptible, perturbations of a correctly classified input can make it misclassified \citep{szegedy2014intriguing}.
Such \emph{adversarial examples} have raised serious concerns as classification failures can entail severe consequences especially in safety-critical applications.

To overcome this problem, two approaches can be taken that complement each other.
The first one is to guide the training process of neural networks so as to improve their robustness. 
This can be accomplished, e.g., by incorporating relaxation methods \citep{dvijotham2018training, wong2018scaling}.
The second approach is to verify the safety of the trained network by providing evidence for the absence of adversarial examples.
For this purpose, heuristic search techniques have been developed that are based on gradient
descent \citep{carlini2017evaluating, szegedy2014intriguing}, evolutionary algorithms \citep{nguyen2015fooled}, or saliency maps \citep{papernot2016limitations}.
However, such methods usually do not provide reliable correctness assertions as they can only show the presence of adversarial examples but never their absence.

For safety-critical applications, however, it is required that robustness properties of neural networks are  rigorously established.
Therefore, techniques based on formal verification have been developed to prove the absence of adversarial examples within a certain distance of a given input.
Ideally, an automated algorithm should either guarantee that this property is satisfied by the network or find concrete counterexamples demonstrating its violation.
The effectiveness and efficiency of such automated approaches crucially depends on how precisely they can estimate the decision boundary of the network.
This is known to be a hard problem for networks with piecewise linear activation functions, such as ReLUs \citep{montufar2014number}.
While the activation function of each node can be decomposed into its linear segments, the number of their possible combinations increases exponentially with the number of nodes.
Therefore, performing the analysis by exhaustively enumerating these combinations is intractable.

The key idea to combat this state-space explosion problem is to apply \emph{over-approximation} by means of symbolic techniques.
Symbolic representations allow to keep track of dependencies across network layers when the actual dependencies become too complex to be represented explicitly.
One of the most promising approaches of this kind is introduced in \citep{wang2018neurify} and implemented in the Neurify tool.
It combines symbolic interval analysis, linear relaxation, and constraint refinement to iteratively minimize the errors introduced during the relaxation process.
As we will see later, however, this approach 
imposes a restriction on the interval bounds established by symbolic relaxation, allowing them to only differ by a constant amount.
In many cases, this entails inaccuracies that could be avoided by obtaining tighter bounds through an independent and, thus, more flexible handling of these bounds.

The contribution of the present paper is a sound technique for decoupling the computation of the upper and lower bounds.
Its key idea is to exploit the fact that when using ReLU activation functions, the value of a node is at least zero.
This can help to improve lower bounds with an otherwise weak estimate, and subsequently also allows to tighten upper bounds. 
Furthermore, we re-implemented part of Neurify, leading to a faster analysis.
By exploiting both implementation-specific improvements and the computation of tighter bounds, our software \enquote{Debona} reduces the runtime of analyses by up to 94\% and increases the rate of successful analyses significantly.
Finally, we provide proofs for tight upper and lower bounds on max-pooling layers in convolutional networks.
We open source Debona to make it publicly available.\footnote{\url{https://github.com/ChristopherBrix/Debona}}

The remainder of this paper is structured as follows.
After giving a brief overview of related work and of preliminaries in Sections~\ref{sec:related_work} and \ref{sec:background}, we detail the mathematical foundations of both Neurify and our own improvements in Sections~\ref{sec:neurify} and \ref{sec:improvements}, respectively.
The setup for assessing the latter and the outcome of the evaluation are described in Section~\ref{sec:experiments}, followed by a short conclusion in Section~\ref{sec:conclusion}.

\section{Related Work}
\label{sec:related_work}

In this section, we focus on symbolic techniques for formal safety analyses of given neural networks, ignoring approaches such as heuristic search algorithms or robustness-oriented training.
Verification methods include constraint solving based on Satisfiability Modulo Theories (SMT) reasoning \citep{dvijotham2018dual, ehlers2017formal, katz2017reluplex, lomuscio2017approach, narodytska2018verifying, pulina2010abstraction, wong2018provable} or Mixed Integer Linear Programming (MILP) solvers \citep{dutta2018output, fischetti2017deep, tjeng2019evaluating}, layer-by-layer exhaustive search \citep{huang2017safety, weng2018fast}, and global optimization \citep{ruan2018reachability}.
Unfortunately, the efficiency of these techniques is usually impaired by the high degree of nonlinearity of the resulting formulae.

To overcome this problem, several linear or convex relaxation methods have been developed to strictly approximate the decision boundary of a network, notably those based on abstract interpretation \citep{li2019analyzing, gehr2018ai2, singh2018fast, singh2019abstract}.
While they tend to scale better than solver-based approaches, this often comes at the price of reduced precision, entailing high false positive rates and problems with identifying real counterexamples that substantiate violations of~safety~properties.

Our approach directly builds on the work described in \citep{wang2018neurify}, which proposes a combined approach that essentially employs symbolic relaxation techniques to identify crucial nodes and that iteratively refines output approximations over these nodes with the help of a linear solver.

While this work mainly focuses on improvements over Neurify, the following alternative verification toolkits exist as well.
\enquote{nnenum} \citep{bak2020vnn}  and \enquote{NNV} \citep{xiang2018tnnls,tran2019formalise,tran2019fm,tran2020cav,tran2020cav_tool} use star sets \citep{bak2017starsets} to propagate the input space through the network.
This can be done either exactly, or using over-approximations.
For the exact case, they also allow to compute the complete counter input set, i.e., all potential adversarial examples.
\enquote{VeriNet} \citep{HenriksenLomuscio20} is similar to Neurify, but extends the verification to networks utilizing sigmoid and tanh activation functions.
\enquote{Oval} \citep{Bunel2017} uses a Branch-and-Bound framework \citep{bunel2020branch} and provides support for GPU based computations.
\enquote{MIPVerify} \citep{Tjeng2019EvaluatingRO} transforms the verification task into a mixed-integer linear programming problem, solvable by third-party toolkits.
\enquote{ERAN} \citep{SinghNIPS:18, singh2019krelu,DeepPoly:19,singhrobustness:19} uses abstract interpretation and extends the analysis to sigmoid, tanh and max-pooling operations.
A comparison of the aforementioned toolkits has been performed in the VNN competition.\footnote{\url{https://sites.google.com/view/vnn20}}
However, as all contestants use different hardware, the results are not easily interpretable.


\section{Notation}
\label{sec:background}
For a given neural network, let the size of the input layer be denoted by $s_0$, followed by $n$ subsequent fully connected feed-forward layers of size $s_1, \ldots, s_n$.
The node values $\hat{x}_1^{(0)}, \ldots, \hat{x}_{s_0}^{(0)}$ of the input nodes represent the network input.
Node inputs in subsequent layers are computed as the weighted sum ${x_{i}^{(l)} = \sum_{j=1}^{s_{l-1}} w_{i,j} \cdot \hat{x}_{j}^{(l-1)}}$.
For all intermediate layers, the node output $\hat{x}_{i}^{(l)}$ is computed by applying a non-linear activation function \mbox{$g$: $\hat{x}_{i}^{(l)} = g(x_{i}^{(l)})$}.
Even though different activation functions exist, this work assumes only ReLU operations are applied, i.e., ${\hat{x}_i^{(l)} = \max\{0, x_i^{(l)}\}}$, as they are easy to compute, piecewise-linear, and commonly used.
The propagation is stopped once the network output  $x_{i}^{(n)}$ is computed.
For notational simplicity, $x_{i}^{(l)}$ and $\hat{x}_{i}^{(l)}$ are referred to as $x$ and $\hat{x}$ whenever the specific $i$ and $l$ are not important.

$Eq^*$ is the higher order function returning the non-linear formula for a given node $x$.
Each node can be approximated by linear upper and lower bounds.
These will be referred to as $Eq_{up}(x)$ and $Eq_{low}(x)$, respectively.
Determining the bounds as a function, as opposed to a simple interval, reduces the overestimation error (see Section~\ref{sec:neurify}).
For the range of valid inputs, specified by the concrete example and its maximal perturbation, both bounds have minimal and maximal values $\underline{Eq_{low}(x)}, \underline{Eq_{up}(x)}$ and $\overline{Eq_{low}(x)}, \overline{Eq_{up}(x)}$, respectively.
Wherever a distinction of upper and lower bound is not necessary for the given argument, as it holds for both instances, they are referred to as $Eq(x)$.
All listed equations can be determined for $\hat{x}$ as well.

\pagebreak

\section{Neurify}
\label{sec:neurify}

For input nodes, upper and lower bounds are determined as a direct result of the given input and the defined $L_{\infty}$ value for possible manipulations.
Other commonly used bounds like $L_1$ and $L_2$ \citep{goodfellow2015adversarial} are supported as well, but not discussed in this paper.
Based on the input bounds, Neurify \citep{wang2018neurify} uses \emph{symbolic propagation} to determine the bounds for each following node.
As opposed to naive interval propagation, symbolic propagation allows to detect common factors of equations, and therefore to tighten the computed bounds:
\begin{align}
a \in [0,1] \Rightarrow a - 0.5a &\mathrel{\overset{\makebox[20pt]{\mbox{\normalfont\tiny\sffamily interval}}}{\in}} [0, 1] - 0.5 \cdot [0, 1] \nonumber \\
& \hspace{-1cm} = [0, 1] - [0, 0.5] = [-0.5, 1] \\
a \in [0,1] \Rightarrow a - 0.5a &\mathrel{\overset{\makebox[20pt]{\mbox{\normalfont\tiny\sffamily symbolic}}}{=}} 0.5a \in [0, 0.5]
\end{align}

Due to the ReLU operations used as the non-linear activation function, the computed symbolic bounds may have to be relaxed to correctly bound the output of each node.
\citet{wang2018neurify} identify three regions the node's bound $Eq(x)$ may fall into:
\begin{enumerate}
	\item $0 \leq \underline{Eq(x)} \leq \overline{Eq(x)}$: If the lowest value taken by the bound is already non-negative, the ReLU operation has no effect on it, and therefore $Eq(\hat{x}) = Eq(x)$.
	\item $\underline{Eq(x)} \leq \overline{Eq(x)} \leq 0$: If the largest value the bound may take is non-positive, the ReLU operation guarantees that $Eq(\hat{x}) = 0$.
	\item $\underline{Eq(x)} \leq 0 \leq \overline{Eq(x)}$: If the boundary indicates that the node may become both negative and positive, there is a non-linear dependence between the node's input and output.
	Thus, the output bounds need to be relaxed.
	\citet{wang2018neurify} refer to these nodes as \emph{overestimated}.
\end{enumerate}

For overestimated nodes, they propose to use \emph{symbolic linear relaxation} to find new bounds for the ReLU output that are valid, but still tight.
They prove that the relaxations
\begin{align}
Eq_{up}(\hat{x}) &= Relax(\max\{0, Eq_{up}(x)\}) \nonumber \\
&\hspace{-1cm}= \frac{\overline{Eq_{up}(x)}}{\overline{Eq_{up}(x)} - \underline{Eq_{up}(x)}} ( Eq_{up}(x) - \underline{Eq_{up}(x)} ) \label{eq:relaxUpper} \\
Eq_{low}(\hat{x}) &= Relax(\max\{0, Eq_{low}(x)\}) \nonumber \\
&\hspace{-1cm}= \frac{\overline{Eq_{low}(x)}}{\overline{Eq_{low}(x)} - \underline{Eq_{low}(x)}} Eq_{low}(x) \label{eq:relaxLower}
\end{align}
minimize the maximal distance between $\max\{0, Eq(x)\}$ and $Eq(\hat{x})$.
This relaxation is visualized in Figures~\ref{fig:scalingUpper} and \ref{fig:scalingLower}.
However, as shown in Section~\ref{sec:scaling}, it is not optimal for approximating $Eq^*(\hat{x})$.

\begin{figure}
	\centering
	\begin{minipage}[t]{.48\textwidth}
		\begin{subfigure}[t]{.5\textwidth}
			\centering
			\begin{tikzpicture}[scale=0.5, transform shape]
			\begin{axis}[grid=none,
			mark = none,
			xmin = 0, ymin = -2.1,
			xmax = 8, ymax = 5,
			axis lines*=middle,
			enlargelimits,
			xtick={0,2,...,16},
			ytick={-6,-4,-2,0,2,...,16},
			tick label style={font=\huge}]
			\addplot[black, domain=0:8, samples=100, line width=2pt]  {0.1*pow(x,2)-1} node [anchor=north east,yshift=-65pt,xshift=10pt] {\huge $Eq^*(x)$};
			\addplot[blue, domain=0:8, samples=100, line width=2pt]  {6.4/8*x-1} node [anchor=north east,yshift=-18pt, xshift=-55pt] {\huge $Eq_{up}(x)$};
			\end{axis}
			\end{tikzpicture}
			\caption{The node input may be negative.}
			\label{fig:scalingUpperA}
		\end{subfigure}%
		\hspace{0.025\textwidth}%
		\begin{subfigure}[t]{.45\textwidth}
			\centering
			\begin{tikzpicture}[scale=0.5, transform shape]
			\begin{axis}[grid=none,
			mark = none,
			xmin = 0, ymin = -2.1,
			xmax = 8, ymax = 5,
			axis lines*=middle,
			enlargelimits,
			xtick={0,2,...,16},
			ytick={-6,-4,-2,0,2,...,16},
			tick label style={font=\huge}]
			\addplot[blue, domain=0:8, samples=100, line width=2pt]  {5.4/(5.4+1)*(6.4/8*x-1+1)} node [anchor=north east,yshift=-10pt, xshift=-55pt] {\huge $Eq_{up}(\hat{x})$};
			\addplot[black, domain=3.16228:8, samples=100, line width=2pt]  {0.1*pow(x,2)-1} node [anchor=north east,yshift=-65pt,xshift=10pt] {\huge $Eq^*(\hat{x})$};
			\addplot[black, domain=0:3.16228, samples=100, line width=2pt]  {0};
			\end{axis}
			\end{tikzpicture}
			\caption{Relaxation ensures positivity of the upper bound.}
		\end{subfigure}
		\setcounter{figure}{0}
		\captionof{figure}{Relaxation of the upper bound is mandatory.}
		\label{fig:scalingUpper}
	\end{minipage}
	\par
	\bigskip
	\begin{minipage}[t]{.48\textwidth}
		\begin{subfigure}[t]{.5\textwidth}
			\centering
			\begin{tikzpicture}[scale=0.5, transform shape]
			\begin{axis}[grid=none,
			mark = none,
			xmin = 0, ymin = -2.1,
			xmax = 8, ymax = 5,
			axis lines*=middle,
			enlargelimits,
			xtick={4,6,...,16},
			ytick={-6,-4,-2,0,2,...,16},
			tick label style={font=\huge},
			extra x ticks={2},
			extra x tick style={
				xticklabel style={yshift=0.5ex, anchor=south}
			}]
			\addplot[black, domain=0:8, samples=100, line width=2pt]  {0.1*pow(x,2)-1} node [anchor=north east,yshift=5pt,xshift=-10pt] {\huge $Eq^*(x)$};
			\addplot[blue, domain=0:8, samples=100, line width=2pt]  {6.4/8*x-2.7} node [anchor=north east,yshift=-90pt, xshift=-35pt] {\huge $Eq_{low}(x)$};
			\end{axis}
			\end{tikzpicture}
			\caption{The node input may be negative.}
			\label{fig:scalingLowerA}
		\end{subfigure}%
		\hspace{0.025\textwidth}%
		\begin{subfigure}[t]{.45\textwidth}
			\centering
			\begin{tikzpicture}[scale=0.5, transform shape]
			\begin{axis}[grid=none,
			mark = none,
			xmin = 0, ymin = -2.1,
			xmax = 8, ymax = 5,
			axis lines*=middle,
			enlargelimits,
			xtick={4,6,...,16},
			ytick={-6,-4,-2,0,2,...,16},
			tick label style={font=\huge},
			extra x ticks={2},
			extra x tick style={
				xticklabel style={yshift=0.5ex, anchor=south}
			}]
			\draw[red!60, line width=4pt] (800,488.9) -- (800,645);
			\draw[green!80, line width=4pt] (0,118.9) -- (0,0);
			\addplot[blue!20, domain=0:8, samples=100, line width=2pt]  {6.4/8*x-2.7};
			\addplot[blue, domain=0:8, samples=100, line width=2pt]  {3.7/(3.7+2.7)*(6.4/8*x-2.7)} node [anchor=north east,yshift=-50pt, xshift=-35pt] {\huge $Eq_{low}(\hat{x})$};
			\addplot[black, domain=3.16228:8, samples=100, line width=2pt]  {0.1*pow(x,2)-1} node [anchor=north east,yshift=5pt,xshift=-10pt] {\huge $Eq^*(\hat{x})$};
			\addplot[black, domain=0:3.16228, samples=100, line width=2pt]  {0};
			\end{axis}
			\end{tikzpicture}
			\caption{Green: improvement of $\underline{Eq_{low}(\hat{x})}$; red: worsening of $\overline{Eq_{low}(\hat{x})}$}
			\label{fig:scalingLowerB}
		\end{subfigure}
		\setcounter{figure}{1}
		\captionof{figure}{Relaxation of the lower bound is optional and a trade off.}
		\label{fig:scalingLower}
	\end{minipage}
\end{figure}

Neurify makes the additional assumption that both bounds are separated only by some scalar $\delta$, i.e.:
\begin{align}
Eq_{up}(x) &= g(x) + \delta_{up}^{x} \\
Eq_{low}(x) &= g(x) + \delta_{low}^{x}
\end{align}
As this implies that the upper and lower bounds cannot be adapted individually, \citet{wang2018neurify} simplify Equations~\ref{eq:relaxUpper} and \ref{eq:relaxLower} by overestimating $\underline{Eq_{up}(x)} \ge \underline{Eq_{low}(x)}$ and $\overline{Eq_{low}(x)} \le \overline{Eq_{up}(x)} $.
This ensures that both bounds are scaled by the same amount, yielding
\begin{align}
Eq_{up}(\hat{x}) &= \frac{\overline{Eq_{up}(x)}}{\overline{Eq_{up}(x)} - \underline{Eq_{low}(x)}} ( Eq_{up}(x) - \underline{Eq_{low}(x)}) \\
Eq_{low}(\hat{x}) &= \frac{\overline{Eq_{up}(x)}}{\overline{Eq_{up}(x)} - \underline{Eq_{low}(x)}}  Eq_{low}(x) \label{eq:wrongNewLower}
\end{align}
However, it introduces an overestimation, implying that the resulting bounds are no longer maximally tight.
As described in Section~\ref{sec:improvements}, it is theoretically possible to determine better estimations of $\underline{Eq_{up}(x)}$, and decoupling the bounds allows to improve the lower bound significantly.

At each node, the upper (lower) bound is determined as the weighted sum of the upper (lower) bound of all previous nodes with a positive weight plus the weighted sum of the lower (upper) bound of all previous nodes with a negative weight.
\begin{align}
Eq_{up}(x_j^{(l+1)}) &= \sum_{i \in [1,\ldots,s_l], w_{j,i} > 0} w_{j,i} \cdot Eq_{up}(\hat{x}_i^{(l)}) \nonumber \\
 & \phantom{= } + \sum_{i \in [1,\ldots,s_l], w_{j,i} < 0} w_{j,i} \cdot Eq_{low}(\hat{x}_i^{(l)})
\end{align}
However, a naive application of this approach leads to weakened bounds.
For $l_i(x) \le f_i(x) \le u_i(x)$ $\forall i \in [1,3]$, $f_2(x) = -\frac{1}{2}f_1(x)$, and $f_3(x) = f_1(x) + f_2(x)$, the bounds of $f_3$ could be computed as
\begin{alignat}{3}
u_3(x) &= u_1(x) + u_2(x) & &= u_1(x) - \frac{1}{2} l_1(x) \\
l_3(x) &= l_1(x) + l_2(x) & &= l_1(x) - \frac{1}{2} u_1(x)
\end{alignat}
even though $f_3(x) = f_1(x) + f_2(x) = f_1(x) - \frac{1}{2} f_1(x) = \frac{1}{2} f_1(x)$ and therefore
\begin{alignat}{3}
u_3(x) &= \frac{1}{2} u_1(x) & &\le u_1(x) - \frac{1}{2} l_1(x) \\
l_3(x) &= \frac{1}{2} l_1(x) & &\ge l_1(x) - \frac{1}{2} u_1(x)
\end{alignat}
Thus, it is important to track different paths that lead to the same node to first simplify the underlying equation as much as possible, before determining the new upper and lower bounds.
Even though this is not described in \citep{wang2018neurify}, it is implemented in Neurify.

After bounds for all nodes have been computed, Neurify uses an LP solver to find a potential adversarial example.
If evaluating the concrete input invalidates it, Neurify splits an overestimated node and performs separate analyses for the assumption that it is either positive, or zero.
Therefore, in both these sub-analyses no overestimation of the given node is necessary, and the bounds are tightened.
\citet{wang2018neurify} split those nodes first that have the highest output gradient, an approach proposed by \citet{wang2018formal}.

\section{Improvements}
\label{sec:improvements}

The proposed improvements over Neurify are two-fold: 
First, we provide Debona~1.0 as a re-implementation of parts of Neurify that eliminates some bugs from the original version.
We note that those bugs prevent Neurify from providing a mathematically sound proof for the (non-)existence of some adversarial examples.
For the inputs analyzed in this work, Neurify wrongly returns \enquote{no adv.\ ex.} for two out of the 7,000 performed analyses, even though an adversarial example provably exists.
Moreover, Debona~1.0 avoids some performance bottlenecks, leading to a significantly faster analysis.
Notably, Debona makes full use of all available computing threads by performing a parallel search over possible splits, whereas Neurify may occasionally not use the full power of parallelization.
Debona also aborts the LP solver after 30 seconds, to avoid long delays due to unfavorable network constraints.
In those situations, the analysis proceeds with the next split.
Those optimizations allow the successful analysis of networks and associated inputs that previously resulted in a timeout or abortion.

Further speedups are realized by determining tighter bounds on the network nodes.
By reducing the overestimation, less splits have to be performed until an adversarial example can be found, or their absence can be proven.
As weak bounds in early layers negatively influence the bounds of later layers, tight approximations are especially important for deep networks.
The major improvement of Debona~1.1 over Neurify is the ability to reduce the approximation error by defining the upper and lower bounds independently of each other.
As described in Section~\ref{sec:neurify}, Neurify assumes that the upper and lower bounds are always parallel to each other.
This implies that the lower bound cannot be changed (other than by moving it up or down by a scalar $\delta$) without influencing the upper bound as well.
For decoupled bounds, we prove the existence of a tighter lower bound, resulting in overall tighter approximations.

In addition to the improvements gained by decoupling the upper and lower bounds, we propose an extension of the analysis that enables the efficient computation of tight bounds for max-pooling layers, under the requirement that each such layer is preceded by a ReLU operation.

\subsection{Zero Bounding}
\label{sec:scaling}
\citet{wang2018neurify} prove that Equation~\ref{eq:relaxLower} minimizes the maximal distance between $\max\{0, Eq_{low}(x)\}$ and ${Eq_{low}}(\hat{x})$.
However, we argue that the lower bound $Eq_{low}(\hat{x})$ should be chosen such that

\begin{align}
\hspace{-10pt} Eq_{low}(\hat{x}) &= \argmin_{Eq_{low}(\hat{x})} \{ ( \underline{\max\{0, Eq_{low}(x)\}} - \underline{Eq_{low}(\hat{x})} ) \nonumber \\
&\phantom{=== } + ( \overline{\max\{0, Eq_{low}(x)\}} - \overline{Eq_{low}(\hat{x})} ) \} \\
&= \argmin_{Eq_{low}(\hat{x})} \{ 0 - \underline{Eq_{low}(\hat{x})} + \overline{Eq_{low}(x)} \nonumber \\
&\phantom{=== } - \overline{Eq_{low}(\hat{x})} \} \\
&= \argmin_{Eq_{low}(\hat{x})} \{ - \underline{Eq_{low}(\hat{x})} - \overline{Eq_{low}(\hat{x})} \label{eq:improvedLower1} \}
\end{align}

where the last transformation is valid as $\overline{Eq_{low}(x)}$ is constant with respect to $Eq_{low}(\hat{x})$.
We highlight that this represents the sum of the maximum error on both the positive and the negative regime of the bound, whereas \citet{wang2018neurify} minimize the maximum of both errors.
By minimizing the sum, we allow the bound estimation to perform a trade off between optimizing both errors, reducing the overall overestimation.

Because the lower bound must be a linear equation, $Eq_{low}(\hat{x}) = m \cdot Eq_{low}(x) + n$.
Therefore
\begin{align}
Eq_{low}(\hat{x}) &= \argmin_{ m \cdot Eq_{low}(x) + n } \{ -\underline{m \cdot Eq_{low}(x) + n} \nonumber \\
&\phantom{===== } - \overline{m \cdot Eq_{low}(x) + n} \} \\
&= \argmax_{ m \cdot Eq_{low}(x) + n } \{ \underline{m \cdot Eq_{low}(x) + n} \nonumber \\
&\phantom{===== } + \overline{m \cdot Eq_{low}(x) + n} \}
\end{align}
In the positive region of the bound, $Eq_{low}(\hat{x})$ must not provide stronger estimates than $Eq_{low}(x)$, i.e., $\restr{Eq_{low}(\hat{x})}{Eq_{low}(x) \ge 0} \le \restr{Eq_{low}(x)}{Eq_{low}(x) \ge 0}$, and both $\underline{Eq_{low}(\hat{x})}$ and $\overline{Eq_{low}(\hat{x})}$ are to be maximized.
Therefore, $n = 0$ and $0 \le m \le 1$. 
It follows
\begin{align}
Eq_{low}(\hat{x}) & \\
&\hspace{-1cm}= \argmax_{ m \cdot Eq_{low}(x) } \{ \underline{m \cdot Eq_{low}(x)} + \overline{m \cdot Eq_{low}(x)} \} \\
&\hspace{-1cm}= \argmax_{ m \cdot Eq_{low}(x) } \{ m \cdot (\underline{Eq_{low}(x)} + \overline{Eq_{low}(x)} ) \} \\
&\hspace{-1cm}= \begin{cases}
Eq_{low}(x), & \text{if } \underline{Eq_{low}(x)} + \overline{Eq_{low}(x)} \ge 0  \\
0, & \text{otherwise }
\end{cases}
\end{align}

\begin{figure}
	\centering
	\begin{minipage}[t]{.48\textwidth}
		\begin{subfigure}[t]{.5\textwidth}
			\centering
			\begin{tikzpicture}[scale=0.5, transform shape]
			\begin{axis}[grid=none,
			mark = none,
			xmin = 0, ymin = -5,
			xmax = 8, ymax = 2,
			axis lines*=middle,
			enlargelimits,
			xtick={0,2,...,16},
			ytick={-6,-4,-2,0,2,...,16},
			tick label style={font=\huge}]
			\addplot[blue, domain=0:8, samples=100, line width=2pt]  {0.75*x-4.5} node [anchor=north east,yshift=-55pt, xshift=-5pt] {\huge $Eq_{low}(x)$};
			\addplot[black, domain=0:8, samples=100, line width=2pt]  {0.0390625*pow(x,2)-1} node [anchor=north east,yshift=10pt,xshift=-30pt] {\huge $Eq^*(x)$};
			\draw (0,500) -- (800,500);
			\draw (0,500) -- (0,0);
			\end{axis}
			\end{tikzpicture}
			\caption{\mbox{$\overline{Eq_{low}(x)} < -\underline{Eq_{low}(x)}$}}
			\label{fig:zeroboundingA}
		\end{subfigure}%
		\hspace{0.025\textwidth}%
		\begin{subfigure}[t]{.45\textwidth}
			\centering
			\begin{tikzpicture}[scale=0.5, transform shape]
			\begin{axis}[grid=none,
			mark = none,
			xmin = 0, ymin = -5,
			xmax = 8, ymax = 2,
			axis lines*=middle,
			enlargelimits,
			xtick={0,2,...,16},
			ytick={-6,-4,-2,0,2,...,16},
			tick label style={font=\huge}]
			\draw[red!80, line width=4pt] (800,500) -- (800,650);
			\draw[green!60, line width=4pt] (0,50) -- (0,500);
			\addplot[blue!20, domain=0:8, samples=100, line width=2pt]  {0.75*x-4.5};
			\addplot[blue, domain=0:8, samples=100, line width=2pt]  {0} node [anchor=north east,yshift=-15pt, xshift=20pt] {\huge $Eq_{low}(\hat{x})$};
			\addplot[black, domain=0:5.05964, line width=2pt] {0};
			\addplot[black, domain=5.05964:8, samples=100, line width=2pt]  {0.0390625*pow(x,2)-1} node [anchor=north east,yshift=10pt,xshift=-30pt] {\huge $Eq^*(\hat{x})$};
			\end{axis}
			\end{tikzpicture}
			\caption{Setting the bound to zero reduces the sum of both errors.}
			\label{fig:zeroboundingB}
		\end{subfigure}
		\setcounter{figure}{2}
		\captionof{figure}{Zero bounding}
		\label{fig:zerobounding}
	\end{minipage}
  \par
  \bigskip		
	\begin{minipage}[t]{.48\textwidth}
		\begin{subfigure}[t]{.5\textwidth}
			\centering
			\begin{tikzpicture}[scale=0.5, transform shape]
			\begin{axis}[grid=none,
			mark = none,
			xmin = 0, ymin = -2,
			xmax = 8, ymax = 5,
			axis lines*=middle,
			enlargelimits,
			xtick={0,2,...,16},
			ytick={-6,-4,-2,0,2,...,16},
			tick label style={font=\huge}]
			\addplot[black, domain=0:8, samples=100, line width=2pt]  {4-0.1*pow(x,2)} node [anchor=north east,yshift=-15pt,xshift=-30pt] {\huge $Eq^*(x)$};
			\addplot[blue, domain=0:8, samples=100, line width=2pt]  {4-(x-2.7)} node [anchor=north east,yshift=80pt, xshift=10pt] {\huge $Eq_{up}(x)$};
			\end{axis}
			\end{tikzpicture}
			\caption{The upper bound is strong on average, but has a large maximum.}
			\label{fig:overapproxA}
		\end{subfigure}%
		\hspace{0.025\textwidth}%
		\begin{subfigure}[t]{.45\textwidth}
			\centering
			\begin{tikzpicture}[scale=0.5, transform shape]
			\begin{axis}[grid=none,
			mark = none,
			xmin = 0, ymin = -2,
			xmax = 8, ymax = 5,
			axis lines*=middle,
			enlargelimits,
			xtick={0,2,...,16},
			ytick={-6,-4,-2,0,2,...,16},
			tick label style={font=\huge}]
			\addplot[black, domain=0:8, samples=100, line width=2pt]  {4-(0.1*pow(x,2))} node [anchor=north east,yshift=-15pt,xshift=-30pt] {\huge $Eq^*(x)$};
			\addplot[blue, domain=0:8, samples=100, line width=2pt]  {4-(0.2*x-0.22)} node [anchor=north east,yshift=30pt, xshift=22pt] {\huge $Eq_{up}(x)$};
			\end{axis}
			\end{tikzpicture}
			\caption{By weakening the bound, the maximum is decreased.}
			\label{fig:overapproxB}
		\end{subfigure}
		\setcounter{figure}{3}
		\captionof{figure}{Minimizing the maximal upper bound}
		\label{fig:overapprox}
	\end{minipage}
\end{figure}

Thus, Debona keeps the lower bound unchanged unless $\underline{Eq_{low}(x)} + \overline{Eq_{low}(x)} < 0$, in which case it is replaced with a constant boundary of zero.
We refer to this process as \emph{zero bounding}.
The tightened lower bound positively impacts both the positive and negative bounds of subsequent layers.
A visualization of zero bounding is shown in Figure~\ref{fig:zerobounding}.

\subsection{Alternative Bound Computation}
As described in Section~\ref{sec:neurify}, the upper bound needs to be relaxed if it could be negative.
To this end, $\underline{Eq_{up}(x)}$ and $\overline{Eq_{up}(x)}$ have to be computed.
While the objective for $Eq_{up}(x)$ is to be as close to $Eq^*(x)$ as possible, there may be other lower bounds that are worse on average, but have a larger minimal or smaller maximal value than $\underline{Eq_{up}(x)}$ and $\overline{Eq_{up}(x)}$.
A visualization of this trade off is shown in Figure~\ref{fig:overapprox}.
In theory, the computation of such alternative bounds could therefore further improve the bound estimation.
However, in practice, we did not find a suitable algorithm without increasing the runtime by unreasonable margins.
We report this idea as a possible research direction for future research.

\subsection{Max-Pooling}
Convolutional networks commonly use max-pooling layers to reduce the layer size, which introduces additional non-linearity.
For a max-pooling operation $\max\{x_{j}^{(l)}, \ldots, x_{j+k}^{(l)}\}$ we propose the following upper and lower bounds:

\begin{align}
& \max\{Eq_{up}(x_j^{(l)}), \ldots, Eq_{up}(x_{j+k}^{(l)})\} \\
&\hspace{20pt} = \max\large\{\sum_{\substack{i \in [1,s_{l-1}],\\ w_{j,i} > 0}} w_{j,i} \cdot Eq_{up}(\hat{x}_i^{(l-1)}) \nonumber \\
&\hspace{20pt} \phantom{= \max\large\{\}} + \sum_{\substack{i \in [1,s_{l-1}],\\ w_{j,i} < 0}} w_{j,i} \cdot \tilde{Eq}_{low}(\hat{x}_i^{(l-1)}), \,\, \ldots, \nonumber \\
&\hspace{20pt} \phantom{= \max\{} \sum_{\substack{i \in [1,s_{l-1}],\\ w_{j+k,i} > 0}} w_{j+k,i} \cdot Eq_{up}(\hat{x}_{i}^{(l-1)}) \nonumber \\
&\hspace{20pt} \phantom{= \max\large\{\}} + \sum_{\substack{i \in [1,s_{l-1}],\\ w_{j+k,i} < 0}} w_{j+k,i} \cdot \tilde{Eq}_{low}(\hat{x}_{i}^{(l-1)}) \large\} \\
&\hspace{20pt} \le \max\large\{\sum_{\substack{i \in [1,s_{l-1}],\\ w_{j,i} > 0}} w_{j,i} \cdot Eq_{up}(\hat{x}_i^{(l-1)}), \,\, \ldots, \nonumber \\
&\hspace{20pt} \phantom{= \max\large\{\}} \sum_{\substack{i \in [1,s_{l-1}],\\ w_{j+k,i} > 0}} w_{j+k,i} \cdot Eq_{up}(\hat{x}_{i}^{(l-1)}) \large\} \\
&\hspace{20pt} \le \sum_{i \in [1, s_{l-1}]} \max\{ w_{j+r,i} \mid r \in [0,k], w_{j+r,i} > 0 \} \nonumber \\
&\hspace{20pt} \phantom{= \max\large\{\}}  \cdot Eq_{up}(\hat{x}_i^{(l-1)}) & \\
&\hspace{20pt} = \sum_{i \in [1, s_{l-1}]} \max\{w_{j,i}, \ldots, w_{j+k,i}, 0\} \nonumber \\
&\hspace{20pt} \phantom{= \max\large\{\}} \cdot Eq_{up}(\hat{x}_i^{(l-1)})
\end{align}
\begin{flalign}
& \min\{Eq_{low}(x_j^{(l)}), \ldots, Eq_{low}(x_{j+k}^{(l)})\} & \\
&\hspace{16pt}  \mathrel{\overset{\makebox[15pt]{\mbox{\normalfont\tiny\sffamily analogous}}}{\ge}} \sum_{i \in [1, s_i]} \min\{w_{j,i}, \ldots, w_{j+k,i}, 0\} \cdot Eq_{up}(\hat{x}_i^{(l-1)}) &
\end{flalign}
where $\tilde{Eq}_{low}(\hat{x}_i^{(l-1)}) = \max\{0, Eq_{low}(\hat{x}_i^{(l-1)})\}$ is a valid tightened lower bound if layer $l-1$ uses ReLUs as the activation function.

The bounds can be further improved by exploiting the following observation.
If for a node $x_j^{(l)}$ it is known that the maximal bound $\overline{x_j^{(l)}}$ is less than the lower bound $\underline{{x}_{k}^{(l)}}$ of a node ${x}_{k}^{(l)}$, $x_j^{(l)}$ is never selected by the max-pooling operation, because ${x}_{k}^{(l)}$ always dominates it.
Therefore, $x_j^{(l)}$ can be omitted from the max-pooling and removed from the above formula, reducing the dependencies of the maximum and minimum computations.
We implement and evaluate this technique in Debona 1.1.

\subsection{Performance Analysis}
As described in Section~\ref{sec:neurify}, it is important to accumulate the effect of each predecessor node over all possible network paths, to avoid weakening the upper and lower bounds.
We note that Neurify is able to do this as part of the forward propagation of bounds, yielding an analysis in ${\mathcal{O}(n \cdot \max_{i \in [0,n]} s_i)}$ steps.
Due to the decoupled bounds, and therefore complicated equation storage, Debona cannot efficiently do so.
Thus, possible paths through the network need to be traced back from the current node to the input layer, increasing the cost of analysis to $\mathcal{O}(n^2 \cdot \max_{i \in [0,n]} s_i)$.
However, as shown in Section~\ref{sec:experiments}, the actual overall runtime is still improved.

\section{Experimental Evaluation}
\label{sec:experiments}

We perform a series of experiments to show both the implementation specific improvements of Debona~1.0 over Neurify and the additional gains drawn from the decoupling of upper and lower bounds as described in Section~\ref{sec:improvements} and implemented in Debona~1.1.

For all three software implementations, we evaluate images on different networks:
\textit{ff2x24}, \textit{ff2x50}, \textit{ff2x512}, \textit{ff3x24}, \textit{ff3x50}, and \textit{ff5x24} are feed-forward networks with the specified number and sizes of hidden layers (i.e., a network structure of e.g.\ $784 \times 24 \times 24 \times 10$ for ff2x24).
\textit{conv} is a convolutional network with two convolutions with 16 and 32 output channels, respectively.
Both convolutions have a window size of 4 and a step size of 2.
\textit{pool} is a convolutional network with one convolution with 16 output channels, followed by a max-pooling operation.
Both have a window size of 4 and a step size of 2.
conv and pool both have two additional fully connected layers of sizes 100 and 10 to reduce the output dimension.
Networks ff2x24, ff2x50, ff2x512 and conv are the same as those used in \citep{wang2018neurify}.
Networks ff3x24, ff3x50, ff5x24 and pool are trained on the MNIST corpus \citep{lecun2010mnist} for six epochs, using Adam \citep{kingma2015adam} with a learning rate of 0.001.
We note that we do not strive for strong network performance, nor do we train the networks to make them more robust against adversarial attacks, as the aim of this work is solely to demonstrate the effectiveness of the analyses.

All experiments are run on a machine with eight cores and 16\,GB RAM.
If the analysis has not terminated after one hour (i.e., no adversarial example has been found, but the network has also not been proven to be robust), it is aborted.
For each experiment, we test for adversarial examples against 1,000 test images from the MNIST dataset for a maximal input perturbation of $L_{\infty} = 10$ for all fully connected networks, $L_{\infty} = 5$ for conv and $L_{\infty} = 1$ for pool.
We report the average wall-clock time spent on a single image, as well as the number of splits performed.
Both values are averaged over the common subset of the 1,000 test images that could be successfully analyzed by Neurify, Debona~1.0 and Debona~1.1.
Images that lead to a timeout or error for one of these implementations were excluded from the average, as the timeout of one hour was arbitrarily chosen, and including them would distort the average.
A complete list of all detailed results is given in the appendix.
All experiments are performed using double precision floating point operations.
We note that the results for Neurify are not completely consistent across repeated executions, as bugs in the implementation may cause race-conditions.
Each verification was run twice and the fastest result was used for the evaluation.

\begin{table*}[t]
	\caption{Performance of the Network Verification for Different Network Structures.}
	\label{table:results}
	\centering
	\begin{tabular}{llrrrrr}
		\toprule
		Network & \multicolumn{1}{l}{Software} & \multicolumn{1}{c}{Time [s]} & Sub-analyses & \multicolumn{3}{c}{Analysis result} \\
		\cmidrule(r){5-7}
		&&&& adv. & \hspace{-5pt} non-adv. & \hspace{-5pt} undetermined \\
		\midrule
		ff2x24 & Neurify & 4.3 \phantom{($-$000\%)} & 140 \phantom{($-$000\%)} & 770 & 222 & 8 \\
		& Debona~1.0 & 1.1 \phantom{0}($-$74\%) & 124 \phantom{0}($-$11\%) & 773 & 227 & 0 \\
		& Debona~1.1 & 0.9 \phantom{0}($-$79\%) & 95 \phantom{0}($-$32\%) & 773 & 227 & 0 \\
		\midrule
		ff2x50 & Neurify & 83.8 \phantom{($-$000\%)} & 2,833 \phantom{($-$000\%)} & 663 & 215 & 123 \\
		& Debona~1.0 & 31.1 \phantom{0}($-$63\%) & 2,728 \phantom{00}($-$4\%) & 677 & 250 & 73 \\
		& Debona~1.1 & 13.1 \phantom{0}($-$84\%) & 1,198 \phantom{0}($-$58\%) & 675 & 264 & 61 \\
		\midrule
		ff2x512 & Neurify & 6.9 \phantom{($-$000\%)} & 135 \phantom{($-$000\%)} & 252 & 12 & 736 \\
		& Debona~1.0 & 3.1 \phantom{0}($-$55\%) & 126 \phantom{00}($-$7\%) & 252 & 17 & 731 \\
		& Debona~1.1 & 0.4 \phantom{0}($-$94\%) & 2 \phantom{0}($-$99\%) & 224 & 59 & 717 \\
		\midrule
		ff3x24 & Neurify & 16.5 \phantom{($-$000\%)} & 457 \phantom{($-$000\%)} & 603 & 371 & 26 \\
		& Debona~1.0 & 5.0 \phantom{0}($-$70\%) & 396 \phantom{0}($-$13\%) & 608 & 391 & 1 \\
		& Debona~1.1 & 3.8 \phantom{0}($-$77\%) & 295 \phantom{0}($-$35\%) & 608 & 391 & 1 \\
		\midrule
		ff3x50 & Neurify & 134.5 \phantom{($-$000\%)} & 3,990 \phantom{($-$000\%)} & 561 & 218 & 221 \\
		& Debona~1.0 & 75.0 \phantom{0}($-$44\%) & 3,894 \phantom{00}($-$2\%) & 562 & 268 & 170 \\
		& Debona~1.1 & 29.6 \phantom{0}($-$78\%) & 1,630 \phantom{0}($-$59\%) & 559 & 301 & 140 \\
		\midrule
		ff5x24 & Neurify & 91.1 \phantom{($-$000\%)} & 2,455 \phantom{($-$000\%)} & 677 & 246 & 77 \\
		& Debona~1.0 & 75.6 \phantom{0}($-$17\%) & 1,984 \phantom{0}($-$19\%) & 690 & 260 & 50 \\
		& Debona~1.1 & 42.8 \phantom{0}($-$53\%) & 1,124 \phantom{0}($-$54\%) & 691 & 270 & 39 \\
		\midrule
		conv & Neurify & 2.8 \phantom{($-$000\%)} &  1 \phantom{($-$000\%)} & 37 & 673 & 290 \\
		& Debona~1.0 & 5.9 ($+$111\%) & 2 ($+$100\%) & 37 & 674 & 289 \\
		& Debona 1.1 & 5.2 \phantom{0}($+86\%$) & 1 \phantom{00}($\pm 0\%$) & 37 & 865 & 98 \\
		\midrule
		pool & Debona 1.1 & 3.8 \phantom{($-$000\%)} & 1 \phantom{($-$000\%)} & 7 & 147 & 846 \\
		\bottomrule
	\end{tabular}
\end{table*}

The results in Table~\ref{table:results} show significant gains both by the implementation specific improvements in Debona~1.0, and the additionally tightened bounds in Debona~1.1.
Across all fully connected networks, the average runtime decreases by 17--74\% when switching to Debona~1.0, demonstrating the importance of highly efficient code.
Even though our optimizations enable maximal parallelism and reduce other bottlenecks, we propose to further optimize the implementation for additional gains.
By open sourcing Debona, we hope to stimulate such development.

Debona~1.1 provides an additional 18--87\% reduction in runtime.
This improvement is solely based on our proposed technique, zero bounding, enabled by the decoupling of the computation of the upper and lower bounds.
With a combined decrease in runtime of 53--94\%, Debona~1.1 allows to test for adversarial examples up to 16 times as fast as the previous state-of-the-art software Neurify.
Furthermore, it enables the analysis of networks and inputs that were previously too complex.
While the decrease in unsuccessful analyses from Neurify to Debona~1.0 is based both on our removal of bugs that cause the analysis to fail or miss critical regions of the search space, and the general speedup, Debona~1.1 provides tighter upper and lower estimations for each node, thus significantly reducing the complexity of the search.
An overview of the bound tightness is given in the appendix.

For the convolutional network, Debona takes about twice as long as Neurify.
The slowdown is caused by additional preprocessing of the network.
However, as the absolute evaluation time is very small, this increase does not significantly reduce the usability of the toolkit.
Because it manages to verify the network robustness for 192 additional inputs, we argue that it is actually superior to the slightly faster but less powerful implementation in Neurify.

As Neurify cannot verify networks that contain max-pooling operations, we only report the results for Debona~1.1.
Even though the analysis fails for many of the inputs, we argue that the proposed bounds are a good first attempt at verifying max-pooling operations, and future research should look into possible extensions to further tighten the bounds and speedup the analysis.

We highlight that Debona~1.1 performs better for the detection of proofs for the non-existence of adversarial examples than for finding specific non-safe instances.
For networks ff2x512 and ff3x50, the number of found adversarial examples decreases, while the number of provenly safe inputs increases significantly.
This indicates that while the tightened bounds reduce the search space sufficiently to prove that many networks are robust to adversarial attacks, for some inputs adversarial examples exist but are hidden in a large search space.
As the improved boundaries influence the search pattern, previously reachable instances may be moved back too far, causing a timeout.
We propose to further investigate this effect, to find search patterns that are robust to changes in the network boundaries and to speed up the detection of adversarial examples.

A fair comparison of Neurify and Debona with other toolkits is not easily possible.
The only recent attempt at a shared benchmark was performed in the VNN competition.
However, as the contestants used vastly different hardware, the results are not comparable.
We tested Debona on four of the settings. For three (trained on MNIST with $L_{\infty} = 25.5$ and \mbox{CIFAR10} \citep{krizhevsky2009cifar} with $L_{\infty} \in \{2, 8\}$), Debona is able to analyze a total of 186 of 238 inputs, which is similar to other top-performing toolkits.
In a fourth setting (trained on MNIST with $L_{\infty} = 76.5$), Debona failed for most inputs due to the large input search space.
We propose to perform a detailed comparison of the different network verification toolkits in future work, ensuring their fair comparison by utilizing standardized hardware.

\section{Conclusion}
\label{sec:conclusion}
In conclusion, we show that by decoupling the computation of the upper and lower bound, significant improvements to the network boundary estimation can be realized.
We prove that zero bounding allows for overall tighter approximations of the lower bound by jointly optimizing its error in both the negative and positive regime.
In our open source implementation, we gain a runtime reduction of up to 94\% by applying this technique in combination with implementation-specific modifications, compared with the state-of-the-art software Neurify.
For convolutional networks, we demonstrate that even though the runtime increases slightly, many additional networks can be verified.
Our proposed bounds for max-pooling operations allow at least some verifications to terminate successfully, motivating future research.

\bibliography{main}

\pagebreak

\begin{appendices}

\subsection*{Tightness of Initial Output Bounds}
Table~\ref{table:tightness} lists the average distance between the upper and lower bounds of the output nodes, averaged over all 1,000 test images and all 10 output nodes.
Neurify and Debona~1.0 have the exact same average distance, as Debona~1.0 is only a re-implementation of Neurify, and none of the removed bugs influenced the bounds that are computed before any splitting is performed.
	
\subsection*{Performance of Analyses Averaged Over All Results}  
Similar to Table~1, Table~\ref{table:allResults} provides the average runtime of each software version over the different network architectures.
However, Table~\ref{table:allResults} reports the average over all analyses that returned a result, not only the subset that was analyzable by all implementations.
The increase in runtime and number of sub-analyses for Debona~1.0 and 1.1 is to be expected, as they include analyses that result in a timeout for previous software versions.
\\
\\

\begin{table}[!h]
	\caption{Tightness of Initial Output Bounds}
	\label{table:tightness}
	\centering
	\begin{tabular}{llr}
		\toprule
		Network & \multicolumn{1}{l}{Software} & Average bound distance \\
		\midrule
		ff2x24 & Neurify & 21.6  \\
		& Debona~1.0 & 21.6  \\
		& Debona~1.1 & 19.1   \\
		\midrule
		ff2x50 & Neurify & 40.5  \\
		& Debona~1.0 & 40.5  \\
		& Debona~1.1 & 34.0 \\
		\midrule
		ff2x512 & Neurify & 227.6   \\
		& Debona~1.0 & 227.6 \\
		& Debona~1.1 & 171.5 \\
		\midrule
		ff3x24 & Neurify & 12.5 \\
		& Debona~1.0 & 12.5 \\
		& Debona~1.1 & 11.0  \\
		\midrule
		ff3x50 & Neurify & 22.4  \\
		& Debona~1.0 & 22.4 \\
		& Debona~1.1 & 17.9 \\
		\midrule
		ff5x24 & Neurify & 24.5  \\
		& Debona~1.0 & 24.5 \\
		& Debona~1.1 & 21.9 \\	
		\midrule
		conv & Neurify & 21.5 \\
		& Debona~1.0 & 21.5 \\
		& Debona~1.1 & 14.6 \\
		\midrule
		pool & Debona~1.1 & 18.9 \\
		\bottomrule
	\end{tabular}
\end{table}

\begin{table*}[]
	\caption{Performance of Analyses Averaged Over All Results}
	\label{table:allResults}
	\centering
	\begin{tabular}{llrrrrr}
		\toprule
		Network & \multicolumn{1}{l}{Software} & \multicolumn{1}{c}{Time [s]} & Sub-analyses & \multicolumn{3}{c}{Analysis result} \\
		\cmidrule(r){5-7}
		&&&& adv. & \hspace{-5pt} non-adv. & \hspace{-5pt} undetermined \\
		\midrule
		ff2x24 & Neurify & 4.3 & 140 & 770 & 222 & 8  \\
		& Debona~1.0 & 1.3  & 150  & 773 & 227 & 0  \\
		& Debona~1.1 & 1.0  & 116  & 773 & 227 & 0 \\
		\midrule
		ff2x50 & Neurify & 83.1 & 2,811 & 663 & 215  & 123 \\
		& Debona~1.0 & 113.0  & 10,065 & 677 & 250 & 73\\
		& Debona~1.1 & 94.2  & 8,572  & 675 & 264 & 61\\
		\midrule
		ff2x512 & Neurify & 6.2 & 121 & 252 & 12 & 736 \\
		& Debona~1.0 & 24.7  & 1,125 & 252 & 17 & 731\\
		& Debona~1.1 & 67.7  & 2,998  & 224 & 59 & 717\\
		\midrule
		ff3x24 & Neurify & 16.5 & 457 & 603 & 371 & 26 \\
		& Debona~1.0 & 5.6  & 446  & 608 & 391 & 1\\
		& Debona~1.1 & 4.2  & 327  & 608 & 391 & 1\\
		\midrule
		ff3x50 & Neurify & 133.7 & 3,964 & 561 & 218 & 221 \\
		& Debona~1.0 & 194.4  & 10,216 & 562 & 268 & 170\\
		& Debona~1.1 & 159.9  & 9,019  & 559 & 301 & 140\\
		\midrule
		ff5x24 & Neurify & 94.5 & 2,563 & 677 & 246 & 77 \\
		& Debona~1.0 & 120.7  & 3,124  & 690 & 260 & 50\\
		& Debona~1.1 & 102.3  & 2,788  & 691 & 270 & 39\\	
		\midrule
		conv & Neurify & 2.8 & 1 & 37 & 673 & 290 \\
		& Debona~1.0 & 6.2 & 2 & 37 & 674 & 289 \\
		& Debona~1.1 & 7.3 & 6 & 37 & 865 & 98 \\
		\midrule
		pool & Debona~1.0 & 3.8 & 1 & 7 & 147 & 846 \\
		\bottomrule
	\end{tabular}
\end{table*}

\end{appendices}

\end{document}